\documentclass{sig-alternate}
\usepackage{times}
\usepackage{url}
\usepackage{graphicx}
\usepackage{amsmath}
\usepackage[ruled,noline,slide,noend]{algorithm2e}
\usepackage{tabulary}
\permission{Permission to make digital or hard copies of part or all of this work for personal or classroom use is granted without fee provided that copies are not made or distributed for profit or commercial advantage, and that copies bear this notice and the full citation on the first page. Copyrights for third-party components of this work must be honored. For all other uses, contact the owner/author(s). Copyright is held by the author/owner(s).}
\conferenceinfo{WWW 2015 Companion,}{May 18--22, 2015, Florence, Italy.} 
\copyrightetc{ACM \the\acmcopyr}
\crdata{978-1-4503-3473-0/15/05.\\
http://dx.doi.org/10.1145/2740908.2742751}
\title{Abstractive Meeting Summarization Using\\ Dependency Graph Fusion}

\numberofauthors{3} %  in this sample file, there are a *total*
\author{
\alignauthor
Siddhartha Banerjee\\
       \affaddr{The Pennsylvania State University}\\
       \affaddr{University Park}\\
       \affaddr{PA, USA - 16802}\\
       \email{sub253@ist.psu.edu}
\alignauthor
Prasenjit Mitra \\
\affaddr{Qatar Computing Research Institute}\\
       \affaddr{Tornado Tower, 18th floor}\\
       \affaddr{Doha, Qatar}\\
       \email{pmitra@qf.org.qa}
\alignauthor 
Kazunari Sugiyama\\
\affaddr{National University of Singapore}\\
\affaddr{13 Computing Drive}\\
\affaddr{Singapore 117417}\\
 \email{sugiyama@comp.nus.edu.sg}
}

\begin{document}
\maketitle
\begin{abstract}
Automatic summarization techniques on meeting conversations developed so far have been primarily extractive, %but such summaries are generally hard to read. 
resulting in poor summaries. %Several abstractive summarization techniques have also been proposed; however, they are either template-based or focused on specific aspects (actions, decisions, etc.) of meetings. %and hence require prior knowledge. 
To improve this, we propose an approach to generate abstractive summaries by fusing important content from several utterances. %Even if a classifier predicts the most important utterances accurately for extractive summarization, the readability of such summaries is very low. 
Any meeting is generally comprised of several discussion topic segments. For each topic segment within a meeting conversation, we aim to generate a one sentence summary from the most important utterances using an \textit{integer linear programming}-based sentence fusion approach. Experimental results show that our method can generate more informative summaries than the baselines. 
%In addition, readability assessments by human as well as log-likelihood estimates obtained from the dependency parser show that our generated summaries are 
%fairly readable and well-formed.
\end{abstract} 
\vspace{-3mm}
\category{I.2.7}{Artificial Intelligence}{Natural Language Processing}[Language generation]
%\category{H.3.3}{Information Storage and Retrieval}{Information Search and Retrieval}[Clustering]
\vspace{-2mm}
\keywords{Abstractive meeting summarization; Integer linear programming}
%clustering, document importance, sentence importance, keyphrase extraction, summarization evaluation}
\section{Introduction}\label{Sec:Intro}
\noindent Meeting summarization helps both participants and non-participants by providing a short and concise snapshot of the most important content discussed in the meetings. A recent study revealed that people generally prefer abstractive summaries~\cite{murray2010generating}.
%An example of two sets of utterances from a meeting in the AMI corpus~\cite{carletta2006ami} is provided in the 
%Table~\ref{tab:MeetingAbsExample} shows two sets of utterances from the AMI corpus~\cite{carletta2006ami} along with their corresponding human written summaries. 
Table~\ref{tab:MeetingAbsExample} shows the human-written abstractive summaries along with the human-generated extractive summaries from a meeting transcript.
%Kaz1: Use the active as much as possible
%Only the utterances that are annotated by human annotators as ``important''~\footnote{According to human annotators, these utterances should be a part of the extractive summary} are shown; other intervening utterances have been avoided. 
%This table shows only the utterances that human annotators have taken as important. 
%We have skipped other intervening utterances not included in the extractive summary. 
%As can be seen, 
%Table~\ref{tab:MeetingAbsExample} also shows that
As can be seen, the utterances are highly noisy and contain unnecessary information. %On the contrary, the human abstract is highly readable and precise. 
Even if an extractive summarizer can accurately classify these utterances as ``important'' and present them to a reader, it is hard to read and synthesize information from such utterances. In contrast, human written summaries are compact and readable. %written in non-conversational style. They are more readable than the extractive summaries and preserve the most important information. %The problem that we aim to solve is as follows: \textit{Given a snippet of utterances in a meeting, construct a new sentence that aims to convey all the ``important'' information discussed in the set of utterances.} 
%Kaz1: No need to insert ``~`` just before ``\footnote{...}''
%Kaz1: ``.'' comes just before ``\footnote{...}'' when the footnote comes at the end of a sentence. 
% Journal papers such as ACM TOIS follows this, so please explore it when you have time. 
% I was also pointed out this for my IJDL journal paper when I prepared final version.   
%Previous approaches to abstractive meeting summarization have relied on template-based~\cite{wang2013focused} or word-graph fusion~\cite{mehdad2013abstractive} based methods. 
%The template-based method was applied to the generation of \textit{focused summaries}.\footnote{Focused summary refers to summaries on specific aspects of the meeting such as actions, decisions, etc. and assumes prior knowledge on the summary aspect type.} The word-graph based fusion technique, on the contrary, used an unsupervised approach to fuse a cluster of utterances generated using an entailment graph based approach. However, this method did not take into consideration any grammatical dependencies between the words, %and hence ungrammatical output was produced in multiple cases.  
%resulting in ungrammatical output in several cases. 

%Kaz1: When we write hiphen, we usually use ``--'' not ``-'' 
We propose an automatic way of generating short and concise abstractive summaries of meetings.  
%Our approach takes into consideration grammatical relations between the words. Further, we do not generate any templates. 
Any meeting conversation includes dialogues on several topics. For example, in Table~\ref{tab:MeetingAbsExample}, the participants converse on two topics: \textit{design features} and \textit{selling prices}. Given the most important sentences within a topic segment, our goal is to generate a one-sentence summary from each segment and appending them to form a comprehensive summary of the meeting. Moreover, we also aim to generate summaries that resemble human-written summaries in terms of writing style.
%We first identify the most important set of utterances in a meeting, and then generate a summary that aggregates information from all such utterances. 
%As can be seen from the Table~\ref{tab:MeetingAbsExample}, 
%As shown in Table~\ref{tab:MeetingAbsExample}, the participants discuss different aspects in the two sets -- \textit{features for remote design} and \textit{selling prices}. 
%We find each topic segment in the meeting conversation, the most important utterances and finally use them to generate a novel summary for each segment.
%To generate abstracts, we need to determine the boundaries of where significant topic changes occur so that we can generate a one-sentence summary for each topic segment. 
%Previous work on meeting summarization~\cite{murray2008summarizing} has shown that \textit{lexical cohesion} is an important indicator in topic identification in meetings. We experiment with two different lexical-cohesion based text segmentation algorithms: LCSeg~\cite{galley2003discourse} and unsupervised Bayesian topic segmentation~\cite{eisenstein2008bayesian}. %Both are based on the idea of lexical cohesion. %In each of the generated segments, we identify the most important utterances using supervised training using a number of discourse and content related features. 
\begin{table}[t]
\centering
\small
\caption{\small{Two sets of extractive summaries and gold standard human generated abstractive summaries from a meeting (Set 2 follows Set 1)}.}
    \begin{tabular}{|p{0.45\textwidth}|}
    \hline
		\textbf{Set 1:} \textbf{Human-generated extractive summary}\\
		\hline
D: um as well as uh characters \\
D: um different uh keypad styles and s symbols.\\ 
D: Well right away I'm wondering if there's um th th uh, like with D\_V\_D players, if there are zones.\\ 
A: Cause you have more complicated characters like European languages, then you need more buttons.\\ 
D: I'm thinking the price might appeal to a certain market in one region, whereas in another it'll be different, so\\ 
D: kay trendy probably means something other than just basic\\ 
\hline
\textbf{Abstractive summary:} The team then discussed various features to consider in making the remote. \\
\hline
	\textbf{Set 2:} \textbf{Human-generated extractive summary}\\
\hline
B: Like how much does, you know, a remote control cost.\\ 
B: Well twenty five Euro, I mean that's um that's about like eighteen pounds or something.\\
D: This is this gonna to be like the premium product kinda thing or \\
B: So I don't know how how good a remote control that would get you. Um.\\ 
\hline
\textbf{Abstractive summary:} The project manager talked about the project finances and selling prices. \\
\hline
\end{tabular}

	\label{tab:MeetingAbsExample}
\end{table}
%To identify the most important utterances, we use a supervised learning approach. 

To aggregate the information from multiple utterances, we adapt an existing integer linear programming (ILP) based fusion technique~\cite{filippova2008sentence}. The fusion technique is based on the idea of merging dependency parse trees of the utterances. The trees are merged on the common nodes that are represented by the word and parts-of-speech (POS) combination. Each edge of the merged structure is represented as a variable in the ILP objective function, and the solution will decide whether the edge has to be preserved or discarded. We modify the technique by introducing an anaphora resolution step and also an ambiguity resolver that takes the context of words into account. Further, to solve the ILP, we introduce several constraints, such as desired length of the output, etc. %The final solution of the problem retains  % that 

To the best of our knowledge, our work is the first to address the problems of readability, grammaticality and content selection jointly for meeting summary generation without employing a template-based approach. %Moreover,  using a text segmentation and fusion based approach. %Further, we add an anaphora resolver before fusing utterances. %While we use extractive summarization component  training, our fusion step is unsupervised. %It only requires certain values for computing edge weights, for which any corpora can be used. 
%\textbf{The following paragraph will be changed to include the metrics}
We conduct experiments on the AMI corpus\footnote{\scriptsize\url{http://groups.inf.ed.ac.uk/ami/corpus/}} that consists of meeting transcripts and show that  our best method outperforms extractive model significantly
on ROUGE-2 scores (0.048 vs 0.026).
%Kaz1: Use the active as much as possible. 
%The generated summaries are compared against the manually constructed abstractive summaries using ROUGE scores~\cite{lin2004rouge}. 
%We compare the generated summaries with the manually constructed abstractive summaries using ROUGE scores~\cite{lin2004rouge}. 
%ROUGE-2 and ROUGE-SU4~\cite{lin2004rouge} scores from our abstractive model (0.048 and 0.087) are significantly better than that of the extractive summaries (0.026 and 0.044) as well as the word-graph based abstractive summarization method~\cite{mehdad2013abstractive} (0.041 and 0.079). We also assess readability of the summaries using a human judge, demonstrating that the summaries generated by our method are fairly well-formed.  
 
%\input{Related_work.tex}
\vspace{-3mm}
\section{Proposed Approach}
%The final step in our approach is to combine information from multiple extracted utterances in each segment that are deemed important by the classifier. 
%Kaz1: You described ``Several techniques ...'' but you show only one reference. 
% It's better to show two or three references.  
%Several techniques have been proposed for sentence fusion tasks~\cite{barzilay2005sentence}. However, 
Dependency fusion on meeting data requires %a robust algorithm that can tackle 
an algorithm that is robust for noisy data as %in utterances there are multiple disfluencies. 
utterances often have disfluencies. %In our case, we apply fusion
%Kaz1: ``Apply *** to ...'' not ``on''
Our work applies fusion to all the important utterances within the topic segment to generate the best sub-tree that satisfies the constraints and maximizes the objective function of the optimization problem.  
Anaphora resolution step replaces pronouns with the original nouns in the previous utterance that they refer to in order to increase the chances of merging. Consider the following utterances:

\small{\textit{``so we're designing a new remote control and um''}}

\small{\textit{``Um, as you can see it's supposed to be original''}} 
\normalsize

\noindent Without pronoun resolution, these two utterances cannot be merged. Once we apply anaphora resolution, \textit{it} in the second utterance is modified to \textit{a new remote control} and then both the utterances are fused into a common structure. 
The utterances are parsed using the Stanford dependency parser. Every individual utterance has an explicit ROOT node. We add two dummy nodes in the graph -- the \textit{start} node and the \textit{end} node to ensure defined start and end points of the merged structure. 
%The ROOT nodes from the utterances are all connected to the \textit{start} node and the last word of every utterance is connected to the \textit{end} node. 
The words from the utterances are iteratively added onto the graph. The words that have the same word form and POS tag are assigned to the same nodes. %A word refers to the tuple of \{word, POS\} this point onwards. 
%We use the ambiguity resolver to resolve words where the merging can be ambiguous. 

\noindent{\textbf{Ambiguity resolver.}} %Let us consider 
Suppose that a new word $w_{i}$ that has $k$ ambiguous nodes where it can be mapped to. The $k$ ambiguous nodes are referred to as mappable nodes. For every ambiguous mapping candidate, 
we first find the words to the left and right of the mappable node of the sentences, and then compute the number of words in both the directions that are common to the words in either direction of the word $w_{i}$. 
%We calculate the directed context in both the directions upto a window size of two words. 
Finally, $w_{i}$ is mapped to the node that has the highest directed context. 
%Kaz1: ``link'' is better than ``tie''? I keep ``tie'' here. Please use more relevant one. 
%Kaz1: Ideally, it is preferable to use larger characters in the figure as they are hard to read.  
\begin{figure}[t]
	\centering
 \fbox{\includegraphics[width=0.45\textwidth, height=5cm]{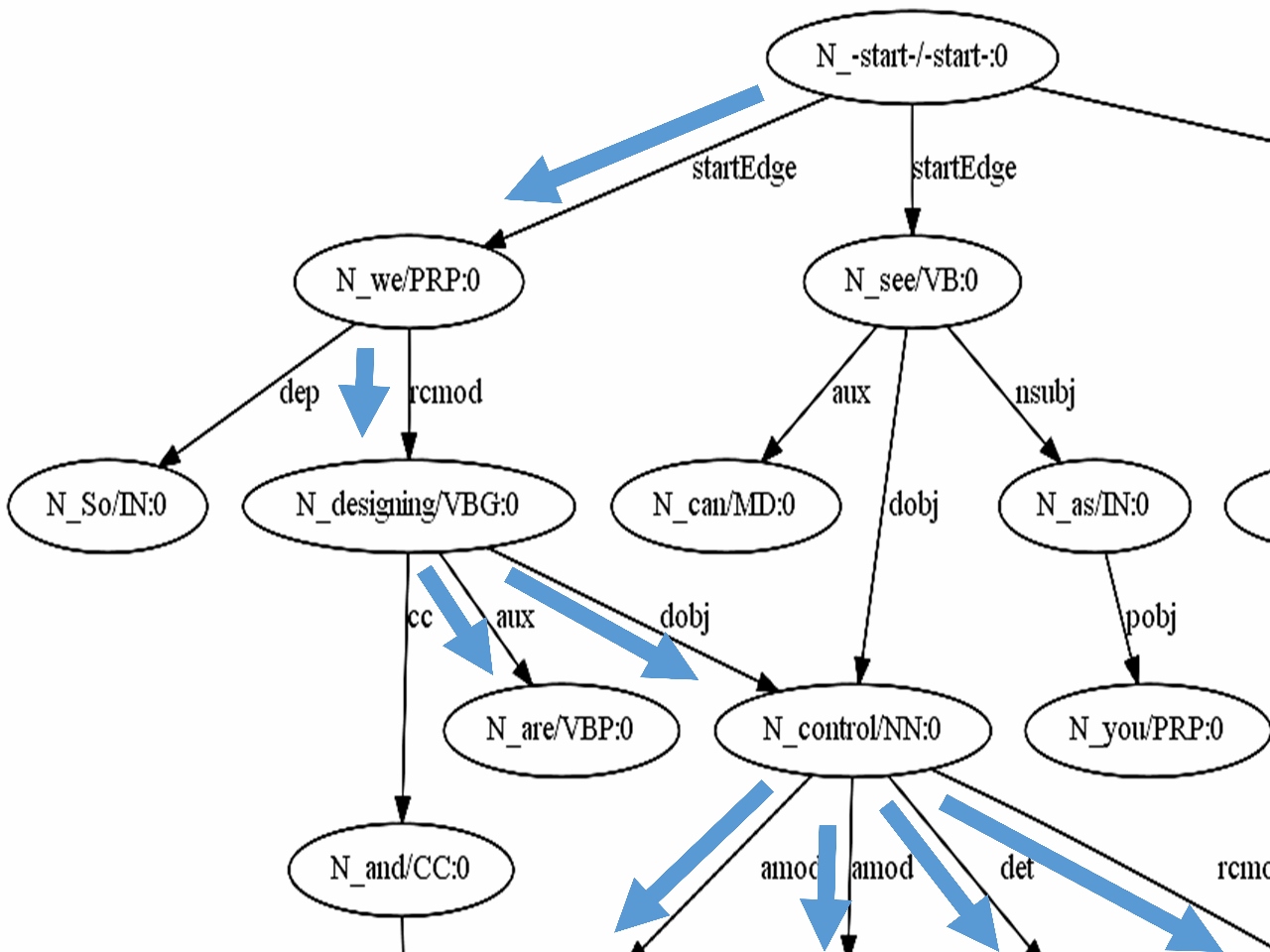}}
	%\fbox{\includegraphics[scale=0.12]{out_changed.png}}
	\caption{\small{A merged dependency graph structure -- edges in blue bold arrows to be retained to generate the summary for each topic segment.}}
	\label{fig:dependencyMerge}
\end{figure}
%\vspace{-1mm}

\noindent{\textbf{ILP formulation.}} Figure~\ref{fig:dependencyMerge} shows the sub-graph (marked using blue bold arrows) that we wish to retain from the merged graph structure to generate a one-sentence summary from several merged utterances. All the sentences generated from each meeting transcript are concatenated to produce the final abstractive summary. We need to maximize the information content of the generated sentence, keeping it grammatical. We model the problem as an integer linear programming (ILP) formulation, similar to the dependency graph fusion as proposed by Fillipova and Strube~\shortcite{filippova2008sentence}. %However, as their dependency tree merging was fundamentally very different from our method, we
%Unlike their method, 
%However, the formulation of our objective function and the constraints are significantly different from their system. 
%we address the constraints in a much different manner. 
%Moreover, they applied it to German language only. 
The directed edges in the graph (binary variables) are represented as $x_{g,d,l}$, where $g$, $d$ and $l$ denote the governor node, dependent node and the label of an edge, respectively. 
%The edges represent the variables in the objective function which can either take value of 1 or 0 depending on whether the edge has to be preserved or deleted.  
We maximize the following objective function:

\small{
\begin{equation}
\sum\nolimits_{x}{x_{g,d,l}\cdot p(l\mid g) \cdot I(d) \cdot \frac{p_{x}}{N} }
\label{eqn:ilp}
\end{equation}
}
\normalsize

As shown in Equation~(\ref{eqn:ilp}), we introduce three different terms: $p(l\mid g)$, $I(d)$ and $\frac{p_{x}}{N}$. %have been introduced in the objective function. 
%which are different from~\cite{filippova2008sentence}. 
Each relation in a dependency graph consists of the governing node, the dependent node and the relation type.
The term $p(l\mid g)$ denotes the probabilities of the labels given a governor node, $g$. %We can calculate these probabilities from any given corpora. 
For every node (word and POS) in the entire corpus, the probabilities are represented as the ratio of the sum of the frequency of a particular label and the sum of the frequencies of all the labels emerging from a node. In this work, we calculate these values using Reuters corpora~\cite{rose2002reuters} to obtain dominant relations from non-conversational style of text. For example, Table~\ref{tab:RelationProbs} shows the probabilities of outgoing edges from a node, (\textit{produced/VBN}). This term assigns the importance of grammatical relations to a node and only the relations that are more dominant from a node will be preferred. %as shown in Table~\ref{tab:RelationProbs}. 
%Kaz1: It is better to show the same style of table as other ones. 
% In addition, I added ``for `\textit{produced/VBN}.''' Please make sure whether it is correct. 
% I also changed the label as the paper displays ``Table 3.3.'' This is fixed after changing the label to ``\label{tab:RelationProbs}''
The term $I(d)$ denotes the informativeness of a node calculated using Hori and Furui's formula~\cite{hori2003new}. 
%In order to compute the informativeness, 
%we used the word significance score from Hori and Furui~\shortcite{hori2003new} with minor modifications. In our case, we denote $I(d)$ as:\\
%we improve the word significance score~\cite{hori2003new} as follows:
%{\vspace{-2mm}
%\begin{equation}
% I(d)=f_{s}\cdot\log\frac{F_{A}}{F_{d}} \label{eq:Informativeness}
%\end{equation}
%}
%In the above formula, 
%In Equation~(\ref{eq:Informativeness}), 
%%$f_{s}$ refers to the frequency of a word in a text segment, $F_{A}$ refers to the sum of the frequencies of all the words in the corpus and $F_{d}$ refers to the frequency of the dependent word $d$ in the entire corpus. 
%$f_{s}$, $F_{A}$, and $F_{d}$ denote the frequency of a word in a text segment, the sum of the frequencies of all the words in the corpus, and the frequency of the dependent word $d$ in the entire corpus (Reuters), respectively. 
%Kaz1: Please specify the Equation ID to make what you refer to clear. 
The last term in Equation~(\ref{eqn:ilp}) is based on the idea of lexical cohesion. Towards the end of any segment, generally, more important discussions might happen that will conclude a particular topic and then start another. In order to take this fact into account, we introduce the term $\frac{p_{x}}{N}$, 
%where $N$ refers to the total number of extracted utterances in a segment and $p_{x}$ refers to the position of the utterance (the edge $x$ belongs to) in the set of $N$ utterances. 
where $N$ and $p_{x}$ denote the total number of extracted utterances in a segment and the position of the utterance (the edge $x$ belongs to) in the set of $N$ utterances, respectively. 

In order to solve the above ILP problem, we impose a number of constraints. 
Some of the constraints have been directly adapted from the original ILP formulation~\cite{filippova2008sentence}. 
% of Fillipova and Strube~\shortcite{filippova2008sentence}. 
For example, we use the same constraints for restricting one incoming edge per node, as well as we impose the connectivity constraint to ensure a connected graph structure. %The other constraints we impose are defined as follows:
%Kaz1: You can write the following equations by using only onee 
% ``\begin{eqnarray} ... \end{eqnarray}'' instead of several ``\begin{equation} ... \end{equation}''
% Maybe, you could be able to save space if you use ``\begin{eqnarray} ... \end{eqnarray}''
% Please try it when you have enough time. 
Further, we restrict the subtree to have just one start edge and one end edge.
This helps in preserving one ROOT node, as well as it limits to one \textit{end} node for the generated subtree. We also limit the generated subtree to have a maximum of 15 nodes that controls the length of the summary sentence. We also add few linguistic constraints that ensure the coherence of the output such as every node can have maximum of one determinant, etc.
%Equation~(\ref{eqn:startend}) limits the subtree to compulsorily have just one start edge and one end edge. This helps in preserving one ROOT node, as well as it limits to one \textit{end} node for the generated subtree. Equation~(\ref{eqn:length}) limits the generated subtree to have a maximum of $\gamma$ nodes. We also add few linguistic constraints that ensure the coherence of the output. For example, every node can have maximum of one determinant, etc.
%The start nodes and end nodes are still a part of the subtree that is generated by solving this optimization problem; hence the value of $\gamma$ needs to be set to 2 more than the desired number of maximum words in the summary sentence. 
%\scriptsize{
%\begin{equation}
%\begin{aligned}
%\forall{l \in startEdge},\sum\nolimits_{l}{x_{g,d,l}}=1, \\
%\forall{l \in endEdge},\sum\nolimits_{l}{x_{g,d,l}}=1
%\end{aligned}
%\label{eqn:startend}
%\end{equation}  
%\begin{equation}
%\sum\nolimits_{x}{x_{g,d,l}}\leq \gamma
%\label{eqn:length}
%\end{equation}
%\begin{equation}
%\sum\nolimits_{g,d}{( x_{g,d,l}+x_{d,g,l} )} \leq 1
%\label{eqn:nocycle}
%\end{equation}
%}
%\normalsize
%Kaz1: It's better to write ``Equation (x)'' than ``constraint (x)'' as the constraints are defined by Equations (3) to (7). 
% I keep ``constraint (x)'' as it is. Please change them if necessary. 
We also impose constraints to prevent cycles in the graph structure, otherwise finding the best path from \textit{start} and \textit{end} nodes might be difficult.
%In order to prevent bidirectional relations between two nodes, we impose constraint~\ref{eqn:nocycle}. 
%To maintain the linguistic quality of the generated sentence, we always include one auxiliary verb (aux), copular verb (cop) and determinant (det) if they exist, using constraint~\ref{eqn:auxcop} and~\ref{eqn:det}. 
%We use the Gurobi software~\cite{gurobi} for the optimization tasks. 
The final graph is linearized to obtain a coherent sentence. In the linearization process, we order the nodes based on their original ordering in the utterance. %Figure~\ref{fig:dependencyMerge} shows an example output from our system that merged several utterances.
\vspace{-2mm}
\section{Experimental Results}
The AMI Meeting corpus contains 20 meeting transcripts in the test set along with their corresponding abstractive (human-written) summaries as well as the annotations of topic segments. 
ROUGE is used to compare content selection of several approaches. % tasks by comparing against human-written summaries. %A Naive Bayes classifier worked best with the set of training features. We also applied a weighted sampling strategy to tackle the imbalance problem in the dataset. 
%We evaluated the performance of the ILP based fusion approach using the segment level extracted utterances as the input. For each segment, we generated a one-sentence summary limiting to a maximum of 20 words. The human-written abstracts, on average, contain close to 300 words.\footnote{We applied the -l 300 parameter while performing ROUGE evaluation to limit summary comparison upto 300 words.} In addition, 
We compared the content selection of our approach to an extractive summarizer~\cite{murray2008summarizing}, which works as a baseline. We also compared our model without using anaphora resolution to see the impact of resolving pronouns. All the summaries were compared against the human-written summaries as reference.
\begin{table}[t]
\scriptsize
	\centering
			\caption{\small{Probabilities of relations from ``\textit{produced/VBN}.''}} \label{tab:RelationProbs}
		\begin{tabulary}{0.48\textwidth}{C|C|C|C|C|C|C}\hline
		auxpass&nsubjpass&aux&prep\_with&agent&prep\_in&advmod\\
		\hline
		0.286&0.214&0.214&0.071&0.071&0.071&0.071\\ \hline	
		\end{tabulary}

\end{table}
\begin{table}[t]
	\centering
	\scriptsize
		\caption{\small{Content selection evaluation. ROUGE scores (R-2 and R-SU4) and log likelihood score (LL) from the Stanford dependency parser.}}
		\begin{tabular}{l|c|c|c}
		\hline
		\textbf{Method} & \textbf{R-2} & \textbf{R-SU4} & \textbf{LL}\\
		\hline
		Our abstractive model& \textbf{0.048} & \textbf{0.087}&-\textbf{125.73}\\
		Our abstractive model \tiny{(without anaphora resolution)}& {0.036} & {0.071}&-130.32\\
		Extractive Model \tiny{(baseline)}& 0.026&0.044&-136.22\\
		\hline
\end{tabular}

	\label{tab:absContentSel}
\end{table}
%As can be seen from the table~\ref{tab:absContentSel}, 
%The results in Table~\ref{tab:absContentSel} show that our method outperforms the other techniques on both ROUGE-2 and ROUGE-SU4 recall scores. Moreover, we performed a coarse estimate of grammaticality using the log-likelihood score from the parser and our technique was better than the extractive method significantly. In future, we plan to design an end-to-end framework for summary generation from meetings.
The results in Table~\ref{tab:absContentSel} show that our method outperforms the other techniques on both ROUGE-2 (R2) and ROUGE-SU4 (R-SU4) recall scores. Moreover, we computed a coarse estimate of grammaticality using the log-likelihood score (LL) from the parser. Our technique significantly outperforms the extractive method. 
In future work, we plan to design an end-to-end framework for summary generation from meetings.

\section*{Acknowledgments}
This material is based upon work supported by the National Science Foundation under Grant No. 0845487.

%\newpage
\vspace{-2mm}
\bibliographystyle{abbrv}
\bibliography{sigproc-sp}  % sigproc.bib is the name of the Bibliography in this case
% ACM needs 'a single self-contained file'!
%\balancecolumns

% That's all folks!
\end{document}